\documentclass[letterpaper, 10 pt, conference]{ieeeconf}  

\IEEEoverridecommandlockouts                              

\pdfoutput=1
\usepackage{graphicx} 
\graphicspath{ {figs/} }
\usepackage{amsmath}
\usepackage{amsfonts}
\usepackage{caption}
\usepackage{subcaption}

\usepackage[dvipsnames]{xcolor}
\usepackage{booktabs} 
\newcommand{\norm}[1]{\left\lVert#1\right\rVert}
\DeclareMathOperator*{\argminB}{argmin}   

\usepackage[normalem]{ulem}
\usepackage{hyperref}
\hypersetup{
    colorlinks=true,
    linkcolor=blue,
    filecolor=magenta,      
    urlcolor=cyan,
}
\usepackage{lmodern}

\title{\LARGE \bf
Optimal Control for Structurally Sparse Systems using Graphical Inference
}

\author{Roshan Pradhan$^{1}$, Shuo Yang$^{1}$, Frank Dellaert$^{3}$, Howie Choset$^{2}$,  and Matthew Travers$^{2}$
\thanks{$^{1}$Roshan Pradhan and Shuo Yang are with the Mechanical Engineering Dept., Carnegie Mellon University, Pittsburgh, PA 15213
        {\tt\small roshannpradhan@gmail.com, shuoyang@andrew.cmu.edu}}%
\thanks{$^{2}$Howie Choset and Matthew Travers are with the Robotics Institute, Carnegie Mellon University, 5000 Forbes Ave, Pittsburgh, PA 15213
        {\tt\small \{choset, mtravers\}@andrew.cmu.edu}}%
\thanks{$^{3}$Frank Dellaert is with the Institute for Robotics and Intelligent Machines, Georgia Institute of Technology, Atlanta, GA 30332 
        {\tt\small fd27@gatech.edu}}
}

\begin{document}

\maketitle
\thispagestyle{empty}
\pagestyle{empty}

\begin{abstract}

Dynamical systems with a distributed yet interconnected structure, like multi-rigid-body robots or large-scale multi-agent systems, introduce valuable sparsity into the system dynamics that can be exploited in an optimal control setting for speeding up computation and improving numerical conditioning. Conventional approaches for solving the Optimal Control Problem (OCP) rarely capitalize on such \textit{structural sparsity}, and hence suffer from a cubic computational complexity growth as the dimensionality of the system scales. In this paper, we present an OCP formulation that relies on graphical models to capture the sparsely-interconnected nature of the system dynamics. Such a representational choice allows the use of contemporary graphical inference algorithms that enable our solver to achieve a linear time complexity in the state and control dimensions as well as the time horizon. We demonstrate the numerical and computational advantages of our approach on a canonical dynamical system in simulation.

\end{abstract}

\section{Introduction} \label{intro}

Many high-dimensional robotic and autonomous systems have inherent \textit{structural sparsity} that is seldom explicitly utilized in real-time optimal control and trajectory optimization. Examples of structural sparsity in dynamical systems include - a multi-rigid-body system like a humanoid robot or a modular serial-link snake robot, where one joint typically constrains at most two adjacent links; or a multi-agent distributed setting like nodes in a wireless mesh network, where communication is restricted to a few neighbouring nodes. Both these classes of systems consist of independent submodules that interact only at the interface defined by local constraints.

Existing optimal control formulations like the Finite-Horizon Discrete-Time Linear Quadratic Regulator (FHDT-LQR) \cite{boyd2004convex}, which are widely used, maintain a linear growth in computational complexity over the time horizon, but suffer from a bottleneck as the system dimensionality scales. Inherently, the FHDT-LQR problem has \textit{temporal sparsity} as can be observed from the banded nature of the Karush-Kuhn-Tucker (KKT) conditions if it is framed as a Quadratic Program (QP) \cite{Laine2019EfficientCO}. Dynamic Programming (DP) approaches exploit this temporal structure to achieve linear complexity in the time horizon. However, it is usually non-trivial to explicitly leverage \textit{structural sparsity}, leading to cubic growth in computation w.r.t problem dimensionality in general. To evade this bottleneck, real-time implementations of optimal control algorithms for high-dimensional robots typically rely on specialized or decoupled formulations of the system dynamics to reduce dimensionality (eg: centroidal dynamics of a humanoid robot).

In this work, we present Sparse Graphical Optimizer (SGOPT) - a general-purpose Optimal Control Problem (OCP) solver for the full-order dynamics of high-dimensional dynamical systems. SGOPT achieves linear growth in complexity w.r.t state and control dimensionsionality as well as time horizon by leveraging the representational power and computational advantages of inference on a graphical model known as a factor graph \cite{Kaess-2017-106169}. In the robotics community, factor graph representations have proved their utility by not only presenting an intuitive interface for modeling real-world, multi-modal estimation problems, but also enabling efficient inference algorithms that exploit sparsity in the model \cite{5979641}. Factor graphs have enabled significant computational improvements that previously inhibited real-time implementations of the full nonlinear Simultaneous Localization and Mapping (SLAM) problem. This was achieved by representing the underlying sparse structure of the system using factor graphs and leveraging efficient inference algorithms to solve large-scale sparse linear systems repeatedly. 

Similar to the SLAM formulation, SGOPT assembles the OCP into a sparse factor graph form, and utilizes the variable elimination (VE) algorithm \cite{Kaess-2017-106169} - a graphical inference tool - to calculate optimal controls based on the cost function. The VE algorithm visits the nodes in the factor graph following a certain ordering, optimizes a local sub-problem, then passes messages to the next node in the ordering. This kind of message passing is made possible due to conditional independence (spatial and temporal) between all but one-hop neighbors in the graph.

In summary, our contributions are - 
\begin{itemize}
    \item We develop an OCP solver based on factor graph inference that achieves linear complexity in state and control dimensions for structurally distributed systems.
    \item We provide comparison experiments to demonstrate how SGOPT improves the computational complexity and numerical conditioning comparing to conventional LQR. 
    \item We motivate extensions of this approach to general multi-rigid-body systems by describing the inner workings of the variable elimination algorithm for a canonical example.
\end{itemize}

\section{Related Work}
Existing optimal control and trajectory optimization solvers already exploit sparsity in different ways. The DP approach to LQR problems can be viewed as a sparse quadratic programming (QP) method \cite{boyd2004convex} as mentioned in Section \ref{intro}. For a state-space system with linear dynamics $\dot{X} = AX+BU$ where $X\in \mathbb{R}^{N_s}$ and $U\in \mathbb{R}^{M}$, the QP formulation may be sparse in the time dimension, but treats the matrices $A$ and $B$ as dense. Some solvers, such as SNOPT \cite{gill2005snopt}, allow users to further specify the sparsity structure of $A$ and $B$, but when the high-dimensional system has arbitrary structure, it is often non-trivial to manually code up the problem-specific sparsity data structure.





In both the robotics and computer graphics communities, researchers have worked on simulation methods for structurally sparse systems that maintain linear complexity (w.r.t number of links) \cite{baraff1996linear, brudigam2020linear}. Since the topology of such systems forms a graphical structure whose adjacency matrix is inherently sparse, linear complexity dynamics simulation can be achieved using sparse matrix operations. When the multi-rigid-body system is a serial chain, the recursive Newton-Euler algorithm computes forward dynamics in $O(n)$ time, where $n$ is the number of the links in the system \cite{featherstone2014rigid}. If the system has a more complicated structure, building a graphical model of the linkage relationship is necessary to exploit spatial sparsity \cite{baraff1996linear}. 
It is shown that this graphical model not only speeds up computation but also improves the simulation precision \cite{brudigam2020linear}.



Recently, there has been a growing interest in leveraging graphical representations for control and motion planning problems. Stochastic optimal control is formulated as a message-passing inference scheme in \cite{10.1145/1553374.1553508}, motion planning is treated as probabilistic graphical inference on factor graphs in \cite{Mukadam_2018} and \cite{dong2016motion}. Xie et al. go a step further and represent the dynamics structure explicitly on a factor graph in \cite{DBLP:journals/corr/abs-1911-10065, xie2020factorgraph, xie2020batch} with demonstrable speedups over sampling-based planners. However, these works rely on stochastic models of robot dynamics with constraint violations incorporated using penalty terms in the objective. For deterministic OCPs, there are instead clear advantages to modeling hard constraints in the optimization both on the estimation and control side \cite{9196649}, and some methods have been proposed to incorporate these into the factor graph framework \cite{5652875, 6842254, Rodrguez2017AFG}. Yang et al. demonstrated that LQR can be modeled quite naturally as inference on a constrained factor graph (where system dynamics are treated as \textit{hard constraints}, i.e. they should be strictly satisfied) \cite{yang2020equality} leading to identical solutions as conventional DP approaches.

\section{Problem Formulation}
\begin{figure}
    \centering
    \includegraphics[width=0.33\textwidth, trim={0cm 0cm 0cm -0.5cm},clip]{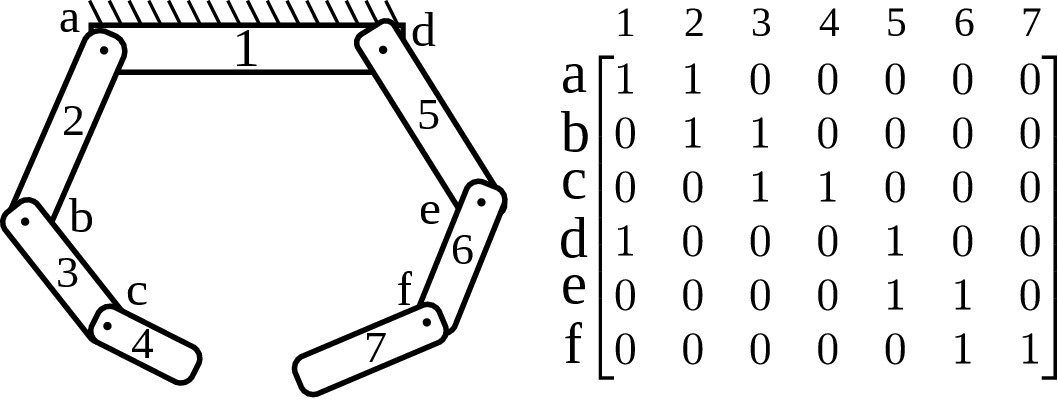}
    \caption{\footnotesize{A planar robot gripper and its sparsity pattern. On the left, numbers indicate links and letters represent joints. On the right, 1s in each matrix row indicate which links connect to the joint represented by the row. The matrix is sparse.}}
    \label{fig:sparsity}
\end{figure}
Our goal is to compute an optimal sequence of states and controls for high dimensional multi-rigid-body systems by first constructing a constrained factor graph (CFG) \cite{5652875} that captures the OCP, and then solving the optimization using VE. We first explain the OCP formulation, present some background into CFGs, and demonstrate how to construct a CFG from the OCP and implement the VE  algorithm. 

\renewcommand{\arraystretch}{0.6}
\begin{figure*} 
\setlength\tabcolsep{1pt}
    \centering
    \begin{subfigure}{0.35\linewidth}
        \centering
        \includegraphics[width = 0.95\linewidth]{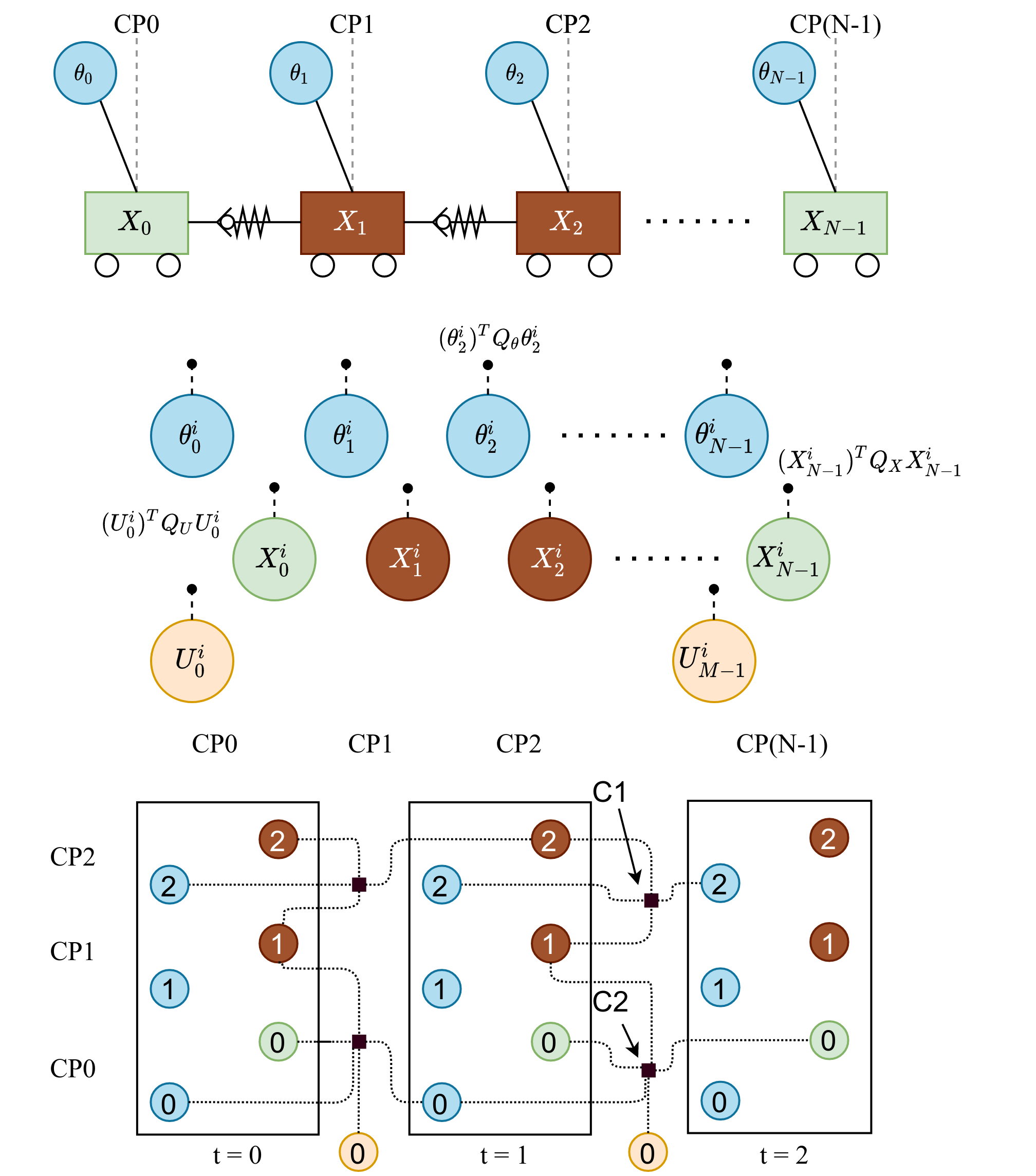}
    \caption{} \label{fig:fg_viz}
    \end{subfigure} 
    \unskip \hspace*{2pt}\vrule~
    \begin{subfigure}{0.55\linewidth} \label{fig:fg_mtx}
        \centering
        \begingroup\makeatletter \def\f@size{5} \check@mathfonts
        $
        \begin{aligned}
        &\begin{array} {c} \hspace{0.1cm} \color{Maroon}{W} \end{array} \quad
        \arraycolsep=2pt\def\arraystretch{1.0} \begin{array} {c c c c c c c c c c c c c c c} 
        \hspace{0.0cm} \color{Maroon}{\theta_0^2} & 
        \hspace{0.1cm} \color{Maroon}{\theta_1^2} & 
        \hspace{0.1cm} \color{Maroon}{X_0^2} & 
        \hspace{0.1cm} \color{Maroon}{X_1^2} & 
        \hspace{-0.1cm} \color{Maroon}{\phantom{-}U_0^1} &
        \hspace{0.0cm} \color{Maroon}{\phantom{-}\theta_0^1} & 
        \hspace{0.0cm} \color{Maroon}{\phantom{-}\theta_1^1} & 
        \hspace{0.0cm} \color{Maroon}{\phantom{-}X_0^1} & 
        \hspace{0.0cm} \color{Maroon}{\phantom{-}X_1^1} & 
        \hspace{0.0cm} \color{Maroon}{\phantom{-}U_0^0} &
        \hspace{0.0cm} \color{Maroon}{\phantom{-}\theta_0^0} & 
        \hspace{0.0cm} \color{Maroon}{\phantom{-}\theta_1^0} & 
        \hspace{0.0cm} \color{Maroon}{\phantom{-}X_0^0} & 
        \hspace{0.0cm} \color{Maroon}{\phantom{-}X_1^0} 
        \end{array} \\
        \normalsize
        &
        \left[  \arraycolsep=1.0pt\def\arraystretch{0.72} \begin{array} {c}
            I \\ I \\ I \\ I \\ I \\ 
            \infty \\ \infty \\
            \infty \\ \infty \\
            I \\ I \\ I \\ I \\ I \\ 
            \infty \\ \infty \\
            \infty \\ \infty \\
            I \\ I \\ I \\ I \\ I 
        \end{array} \right]
        \left[  \arraycolsep=1.7pt\def\arraystretch{0.75}
        \begin{array} {c c c c c c c c c c c c c c | c} 
        Q'_{\theta} & 0  & 0  & 0  & 0   & 0  & 0 & 0 & 0 & 0 & 0 & 0 & 0 & 0 & 0 \\
        0  & Q'_{\theta} & 0  & 0  & 0   & 0  & 0 & 0 & 0 & 0 & 0 & 0 & 0 & 0 & 0 \\
        0  & 0  & Q'_{X} & 0  & 0   & 0  & 0 & 0 & 0 & 0 & 0 & 0 & 0 & 0 & 0 \\
        0  & 0  & 0  & Q'_{X} & 0   & 0  & 0 & 0 & 0 & 0 & 0 & 0 & 0 & 0 & 0 \\
        0  & 0  & 0  & 0  & Q'_U   & 0  & 0 & 0 & 0 & 0 & 0 & 0 & 0 & 0 & 0 \\
        I  & 0  & 0  & 0  & -B_*  & -A_* & 0 & -A_* & -A_* & 0 & 0 & 0 & 0 & 0 & 0 \\
        0  & I  & 0  & 0  &  0  &  0 & -A_* & -A_* & -A_* & 0 & 0 & 0 & 0 & 0 & 0 \\
        0  & 0  & I  & 0  & -B_*  & -A_* &  0 & -A_* &-A_* & 0 & 0 & 0 & 0 & 0 & 0 \\
        0  & 0  & 0  & I  &  0  &  0 & -A_* & -A_* &-A_* & 0 & 0 & 0 & 0 & 0 & 0 \\
        0 & 0 & 0 & 0 & 0 & Q'_\theta & 0  & 0 & 0  & 0 & 0 & 0 & 0 & 0 & 0 \\
        0 & 0 & 0 & 0 & 0 & 0 & Q'_\theta  & 0 & 0  & 0 & 0 & 0 & 0 & 0 & 0 \\
        0 & 0 & 0 & 0 & 0 & 0 & 0  & Q'_X & 0  & 0 & 0 & 0 & 0 & 0 & 0 \\
        0 & 0 & 0 & 0 & 0 & 0 & 0  & 0 & Q'_X  & 0 & 0 & 0 & 0 & 0 & 0 \\
        0 & 0 & 0 & 0 & 0 & 0 & 0  & 0 & 0  & Q'_U & 0 & 0 & 0 & 0 & 0 \\
        0 & 0 & 0 & 0 & 0 & I  & 0  & 0  & 0  & -B_*  & -A_* & 0 & -A_* & -A_* & 0 \\
        0 & 0 & 0 & 0 & 0 & 0  & I  & 0  & 0  & 0  &  0 & -A_* & -A_* &-A_* & 0 \\
        0 & 0 & 0 & 0 & 0 & 0  & 0  & I  & 0  & -B_* & -A &  0 & -A_* &-A_* & 0 \\
        0 & 0 & 0 & 0 & 0 & 0  & 0  & 0  & I  &  0  &  0 & -A_* & -A_* &-A_* & 0 \\
        0 & 0 & 0 & 0 & 0 & 0 & 0 & 0 & 0 & 0 & Q'_\theta & 0  & 0 & 0  & 0 \\
        0 & 0 & 0 & 0 & 0 & 0 & 0 & 0 & 0 & 0 & 0 & Q'_\theta  & 0 & 0  & 0 \\
        0 & 0 & 0 & 0 & 0 & 0 & 0 & 0 & 0 & 0 & 0 & 0  & Q'_X & 0  & 0 \\
        0 & 0 & 0 & 0 & 0 & 0 & 0 & 0 & 0 & 0 & 0 & 0  & 0 & Q'_X  & 0 \\
        \end{array} 
        \right]
        \end{aligned}
        $
        \endgroup
        
    \caption{} \label{fig:fg_mtx}
    \end{subfigure} 
    \caption{
    \footnotesize{
    (a) \textbf{Top}: Schematic of N-CP system, green carts are actuated, brown carts are passive. \newline 
    \textbf{Mid}: Factor graph for the system at single time-step, with expressions for selected \textit{unary cost factors} displayed. \newline 
    \textbf{Bottom}: The Mid factor graph black-box for $N=3, M=1$ rotated by 90 degrees a.c.w, and connected over $T=3$ time-steps. Connections only for selected \vspace{0.3cm}  \textit{dynamics constrained factors} are displayed. \newline 
    (b) Matrix representation of the factor graph for N-CP system with $N=2$, $M=1$, and $T=3$. The left vector is the diagonal of $W$, the right matrix is $[F|g]$. The symbols $Q'_*$ represents the square root of $Q_*$ terms in the cost function (Eqn \ref{pend_ocp}). $A_*$ and $B_*$ represents the state matrices in Eqn \ref{c1_c2}.
    }
    }
    \label{fig:key_figure}
\end{figure*}


\subsection{Optimal Control Problem}

As per the standard formulation, the cost function to be minimized is given by 
\begin{equation}\label{ocp}
\begin{aligned}
\min_{X,U} \quad &X_{T-1}^T Q_{X_f} X_{T-1} + \sum_{i=0}^{T-2}{X_i^T Q_X X_i + U_i^T Q_U U_i}\\
\textrm{s.t.} \quad & X_{i+1} = f(X_i, U_i)\\
\end{aligned}
\end{equation}

\vspace{0.1cm}

Where $X_i \in \mathbb{R}^{N_s}$, $U_i \in \mathbb{R}^{M}$, and $T$ are the state, controls, and time horizon respectively. $Q_X, Q_U, Q_{X_f}$ are weights in the form of diagonal matrices. This problem can be solved with conventional FHDT-LQR, which has a computational complexity of $O(T(N_s+M)^3)$ \cite{boyd2004convex}. 

To improve computational efficiency, we leverage the structural sparsity inherent in the system dynamics. Most practical robots consist of rigid links, whose relative poses constitute the state, and connecting joints impose constraints that are explicitly or implicitly described by the system dynamics $f$. Fig. \ref{fig:sparsity} illustrates the sparsity pattern among links and joints of a planar robot. Instead of explicitly constructing this pattern matrix, we use the factor graph to keep track of the sparsity. 

 In this paper, we present our approach and results on a structurally-sparse canonical system of $N$ identical Cart-Pole (CP) systems in a series, with each cart linked to its neighbor by spring-damper connections. Since the cartpole system is canonical example in control theory, the method we derived on it can generalize to other systems.
 The number ($M$) and order of actuated carts is a user input, but unless otherwise specified they are assumed to be evenly distributed in the CP series according to an actuation ratio $\rho$, where $M=round(\rho N)$. Most of our analysis focuses on linear systems, hence the CP series system is linearized about unstable equilibrium (all pendulums upright). The cost function for this system over $T$ time steps is:
 
  \begin{equation}\label{pend_ocp}
\begin{footnotesize}
\begin{aligned}
\min_{X,\theta,U} &\quad \left[ \arraycolsep=1.0pt\def\arraystretch{1.0} \begin{array}{c c} X_{T-1}^T & \theta_{T-1}^T \end{array} \right]^T 
\left[ \arraycolsep=1.0pt\def\arraystretch{1.0} \begin{array}{c c} Q_{X_f} & 0 \\ 0 & Q_{\theta_f} \end{array} \right] 
\left[ \arraycolsep=1.0pt\def\arraystretch{1.0} \begin{array}{c c} X_{T-1}^T & \theta_{T-1}^T \end{array} \right] + ... \\
& \hspace{10pt} \sum_{i=0}^{T-2}{ \{
\left[ \arraycolsep=1.0pt\def\arraystretch{1.0} \begin{array}{c c} X_i^T & \theta_i^T \end{array} \right]^T 
\left[ \arraycolsep=1.0pt\def\arraystretch{1.0} \begin{array}{c c} Q_{X} & 0 \\ 0 & Q_{\theta} \end{array} \right] 
\left[ \arraycolsep=1.0pt\def\arraystretch{1.0} \begin{array}{c c} X_i^T & \theta_i^T \end{array} \right]
+ U_i^T Q_U U_i \} }\\
\textrm{s.t.} & \quad \left[ \arraycolsep=1.0pt\def\arraystretch{1.0} \begin{array}{c} X_{i+1} \\ \theta_{i+1} \end{array} \right] = f(X_i, \theta_i, U_i)\\
\end{aligned}
\end{footnotesize}
\end{equation} 
\vspace{0.1cm}

Where, $X_i \in  \rm I\!R^{2N x 1}$ denotes position and velocities of the carts, $\theta_i \in  \rm I\!R^{2N x 1}$ denotes angles and angular velocities of the carts, $U_i \in  \rm I\!R^{M x 1}$ denotes forces on actuated carts. When we look at the CFG construction in the next section, we will take subsets from $X_i$, $\theta_i$, $U_i$, and $f$ to describe local state transitions between one-hop neighbors.




%

\subsection{Constrained Factor Graph}
In this section, we transform the OCP for the N-CP system into a CFG. The transformation is similar to that described in \cite{yang2020equality}. In this work, we further subdivide the states and controls of the dynamical system into factor graph nodes instead of treating the full state as one single node. 


A factor graph can be interpreted in different views. In robot state estimation \cite{Kaess-2017-106169}, it is commonly viewed as a joint probability distribution over a set of variables. Solving a factor graph inference means finding the Maximum A-Posteriori (MAP) distribution of variables. The factor graph data structure makes it easy to calculate marginalized conditional probabilities of variables. In another view, a factor graph is just a sparse weighted least square problem \cite{yang2020equality}, where graph nodes are decision variables and graph edges form the matrix for the least square problem. 

In this paper we will take the weighted least square view since it is easier to draw connections to sparse matrix factorization. A weighted least square problem can be formulated as 
\begin{align}
    \min_{Y} \ (FY-g)^TW(FY-g) \label{eqn:wls}
\end{align}
where $W=\text{diag}(w_1,w_2,\dots)$ is a diagonal matrix with non-negative weights. Alternatively we can write the problem as $\norm{FY - g}^2_{\bf{\Sigma}}$ using Mahalanobis norm in which $\Sigma=W^{-1}$. The matrices $F$, $g$ and $W$ can be encoded in factor graph form as shown in Fig \ref{fig:key_figure}. If some weights in $W$ are infinity, then the problem becomes a constrained weighted least square, and the corresponding graph is a constrained factor graph.  
 
Nodes of a factor graph contain either the physical state of a rigid link or a control force input. We denote $X_j^{i} \in  \rm I\!R^{2 x 1}$ as the position and velocity of the cart and $\theta_j^i \in  \rm I\!R^{2 x 1}$ as the angle and angular velocity of the pendulum in CP $j$. $U_j^{i} \in  \rm I\!R$ refers to the $j$th force acting on the associated cart. We define $X_i = vec(X_0^{i}\ X_1^{i}\ X_2^{i} \dots)$ and $U_i = vec(U_0^{i}\ U_1^{i} \dots)$, where $vec$ stacks vectors vertically. We write the value of all nodes in the graph as $Y_i=vec(X_i,U_i)$. For time step $i$, we lay down nodes as shown in the Mid part of Fig. \ref{fig:fg_viz}, where the location of nodes align with the topology of the actual system. 

The edges, or factors, in a factor graph represent either unconstrained factors (finite weight) or constrained ones (infinite weight). The cost function from Eqn \ref{pend_ocp} is represented in the factor graph as unconstrained unary factors connected to the respective variables. For a  diagonal $Q_X$ matrix, we define a cost factor $\norm{Q_X^{\frac{1}{2}} X_j^i}^2_\Sigma$ where $\Sigma = I$. The factor occupies one block row of $F$ and $g$ with $Q_X^{\frac{1}{2}}$ located at the colum corresponding to variable $X_j^i$ in $F$. Similarly, we can define cost factors for $\theta_j^i$ and $U_j^i$. In our example graph in Fig \ref{fig:fg_viz} \textit{Bottom}, we have $N$=3 , $M$=1 acting on the first cart, and $T=2$. After linearizing about current operating state, the expressions for constrained factors labeled \textit{C1} and \textit{C2} are:


\begin{equation} \label{c1_c2}
\resizebox{.44 \textwidth}{!} 
{
$
\begin{aligned}
&C1: \ \norm{ \theta^2_2 - A_{\theta\_\theta}\theta^1_2 - A_{\theta\_X} X^1_2 - A_{\theta\_nbX} X^1_1  }^2_{\Sigma_0}  \\
&C2: \ \norm{X^2_0 - A_{X\_X}X^1_0 - A_{X\_nbX}X^1_1 - A_{X\_\theta}\theta^1_0 - B_X U^1_0 }^2_{\Sigma_0}
\end{aligned}
$
}
\end{equation}

\vspace{0.2cm} 

Where $\Sigma_0=\mathbf{0}$ (infinite weight). Here $A_{\alpha\_\beta}$ with $\alpha, \beta \in \{X,\theta\}$   represents the local linear discrete dynamics state matrix from node type $\beta$ to node type $\alpha$ (at next time-step). Similarly $A_{\alpha\_nb\beta}$ represents the state matrix from $\beta$ to $\alpha$ for neighboring CPs. $B_\alpha$ is the input matrix for node type $\alpha$ due to actuator force. For example: \vspace{0.3cm} 

$A_{\theta\_nbX} = \left[ \arraycolsep=1.0pt\def\arraystretch{1.0}  
\begin{array}{c c} 
\frac{k \Delta t^2}{2 m_c L}  & \frac{c \Delta t^2}{2m_cL} \\ 
\frac{k\Delta t}{m_c L} & \frac{c \Delta t}{m_c L} \end{array} \right],
A_{X\_nb\theta} = \left[  \arraycolsep=1.0pt\def\arraystretch{1.0} 
\begin{array}{c c} 
\frac{m_p g \Delta t^2}{2 m_c}  & 0 \\ 
\frac{m_p g \Delta t}{m_c} & 0 \end{array} \right]$

\vspace{0.3cm} 

Where $k,c$ are spring-damper properties, $m_c,m_p$ are cart and pendulum masses respectively, $L$ is pendulum length, $\Delta t$ is the time discretization.

A complete representation in both factor graph form  and matrix form for the N-CP system is shown in Fig. \ref{fig:key_figure}. The matrix $F$ is visibly sparse, and the sparsity pattern is explicitly captured in the factor graph structure. While we present a specific example of a N-CP system, this methodology can be just as easily applied to any multi-rigid-body system.



\section{Technical Approach}
We use the Variable Elimination algorithm \cite{Kaess-2017-106169} to solve Problem \ref{eqn:wls}. For the N-CP system, the dimension of the decision variable vector $Y$ is $4N*T + M*(T-1)$. By leveraging the sparsity of the graph, we can achieve $O((M+N_s)T)$ solution complexity, where $N_s=4N$. 
\subsection{Variable Elimination on CFG}\label{ve_details}
\begin{figure} [!h]
  \centering
 \includegraphics[width = 0.9\linewidth]{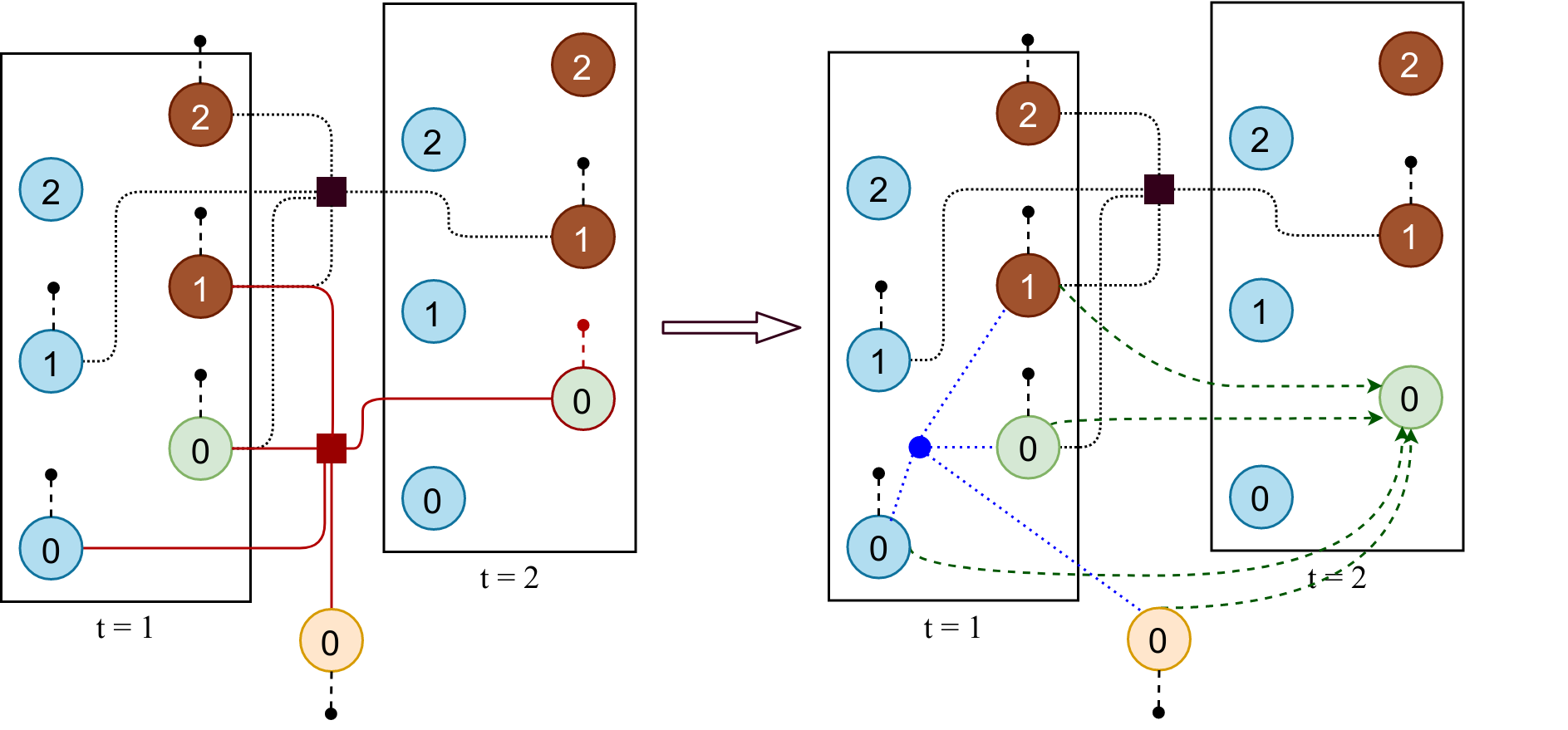}
  \caption{\footnotesize{VE step 1 - Eliminate cart node 0 at $t=2$}}
  \label{fig:ve_1}
\end{figure}

The VE algorithm visits nodes in the factor graph following an \textit{elimination order} and updates different parts of $F$ and $g$ in Eq. \ref{eqn:wls}. After each node is visited once, $F$ will be converted into a sparse upper triangular matrix so $Y$ can be calculated in linear time. For the sake of demonstration, we choose an intuitive ordering of the variable elimination (cart node\textrightarrow pendulum node\textrightarrow control node) backwards in time from $t=T-1$ to $t=0$. This ordering is easy to follow, and makes it possible to extract Markovian feedback policies along with the optimized controls. Section \ref{ordering} will treat the choice of variable ordering further. The square factors in Fig \ref{fig:ve_1} represent hard-constrained factors, while circular factors represent unconstrained factors with a weight $W=\Sigma^{-1}$.

\begin{figure} [!h]
  \centering
  \includegraphics[width = 0.9\linewidth, trim={0cm 0cm 0cm -2cm},clip]{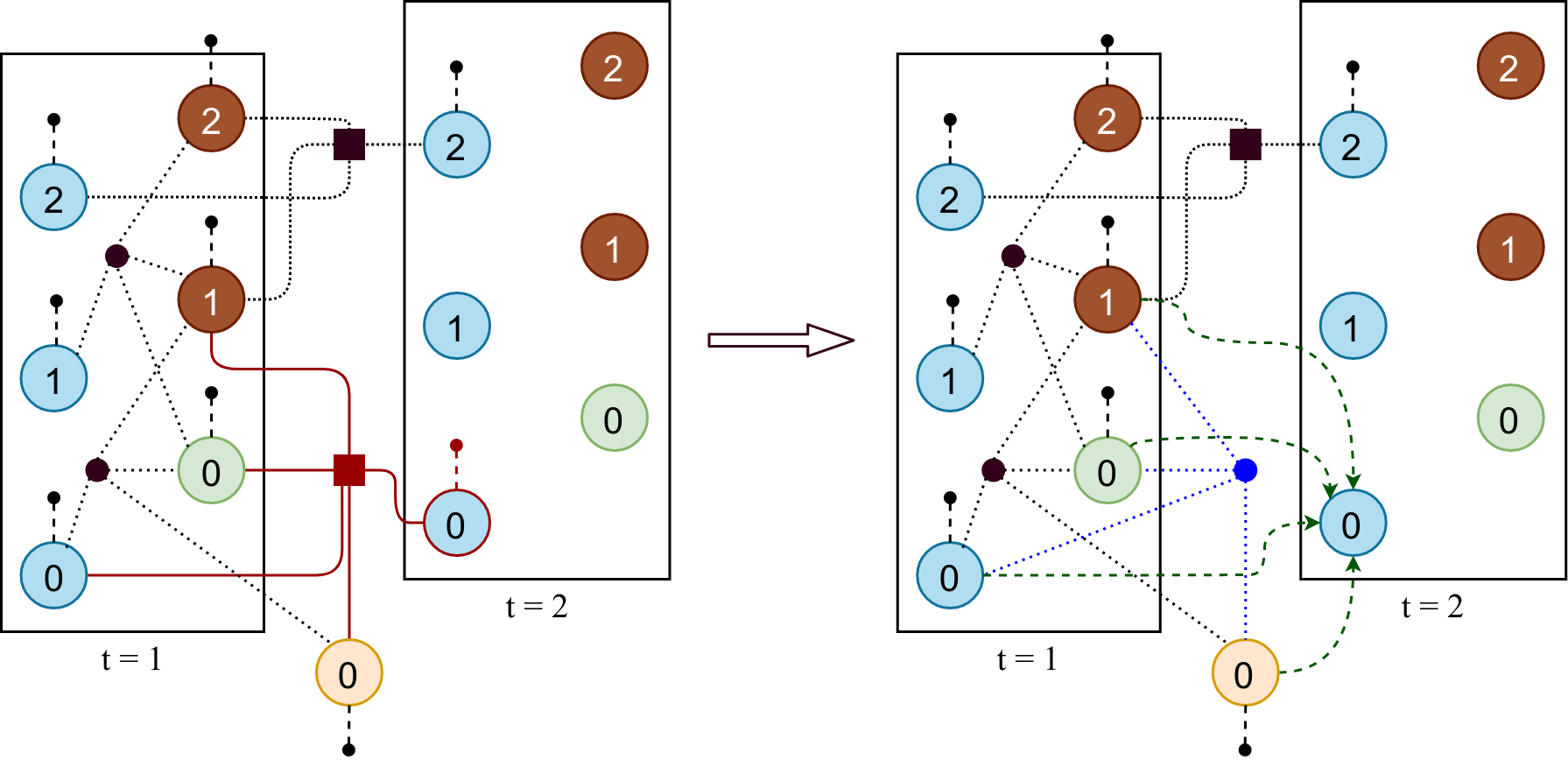}
  \caption{\footnotesize{VE step 2 - Eliminate node pendulum 0 at $t=2$}}
  \label{fig:ve_2}
\end{figure}

\begin{figure} [!h]
  \centering
  \includegraphics[width = 0.9\linewidth]{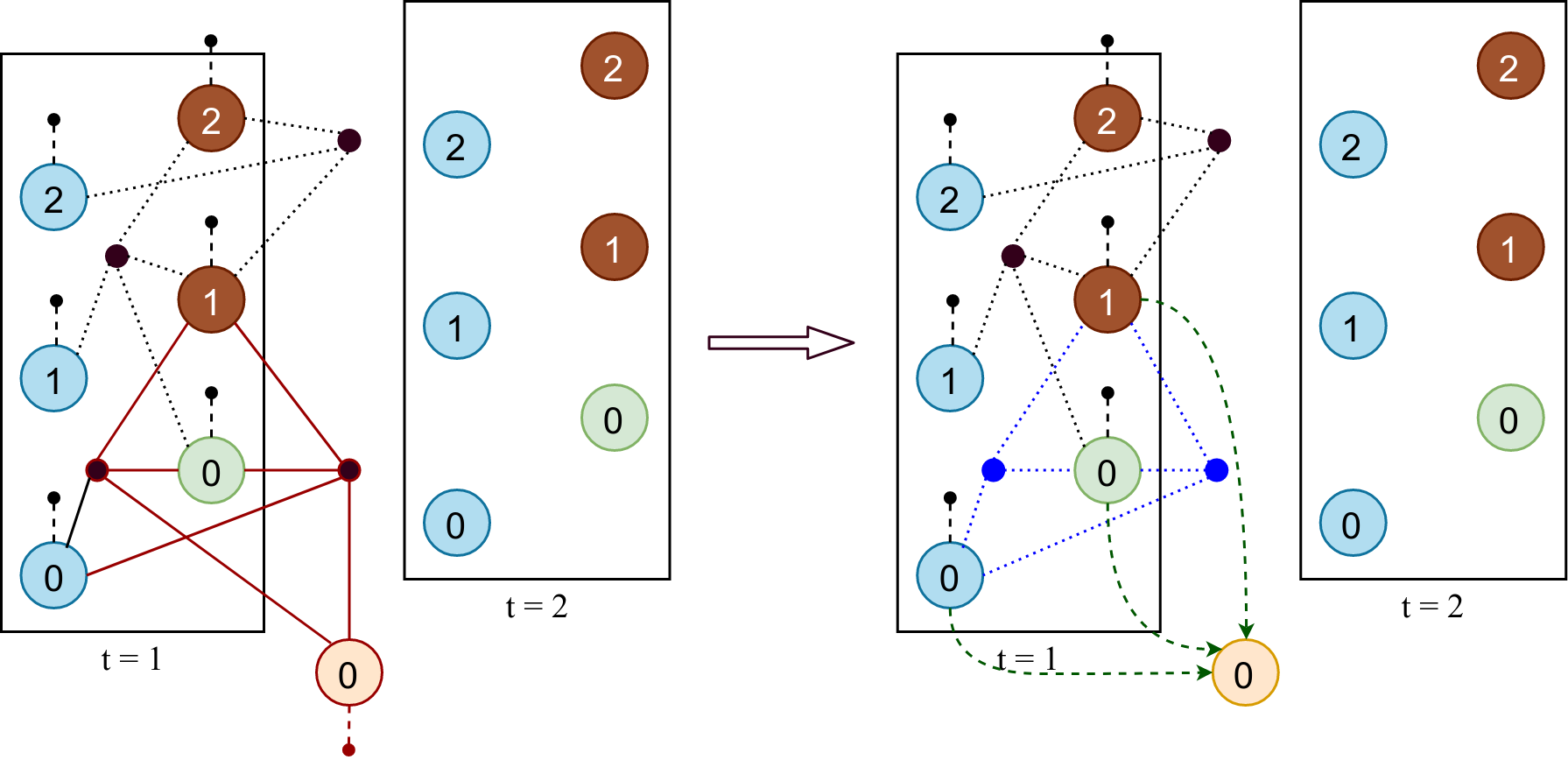}
  \caption{\footnotesize{VE step 3 - Eliminate node control 0 at t = 1}}
  \label{fig:ve_3}
\end{figure}

For each variable $Y_k$ in the elimination order, we construct a local sub-problem. We first identify the separator $S_k$: the set of other nodes sharing factors with $Y_k$. We then extract sub-matrices $F_k$, $W_k$, $g_k$ that are blocks of $F$, $g$, $W$ corresponding to the node variable and its separator. These matrices form a local least-squares problem, which is solved using a Modified Gram Schmidt (MGS) orthogonalization \cite{gulliksson1995modified} in the presence of hard-constrained factors ($\Sigma = \mathbf{0}$) and standard unconstrained QR Factorization otherwise. For the sake of visual clarity, only selected factors and arrows are shown in Figs \ref{fig:ve_1}-\ref{fig:ve_3}.

\textbf{Eliminate a cart node $X_0^2$:} Node $X_0^2$ has a cost factor and constraint factor highlighted in red, with the separator containing $\{X_0^1, X_1^1, \theta_0^1, U_0^1\}$.Eqn \ref{ve_1_eqn} shows the local constrained least squares problem and Eqn \ref{ve_1_matrix} shows the corresponding MGS orthogonalization on $[F_k | g_k]$.

\vspace{-0.65cm}
\begin{equation}   \label{ve_1_eqn}
\resizebox{.4 \textwidth}{!} 
{
$
\begin{aligned}
&X_0^{2^*} = \argminB_{X_0^2} \textcolor{red}{(X_0^2)^T Q_X X_0^2} \\
&\textrm{s.t.} \hspace{6pt} \textcolor{red}{ X^2_0 - A_{X\_X}X^1_0 - A_{X\_nbX}X^1_1 - A_{X\_\theta}\theta^1_0 - BU^1_0 = 0} \\
&\hspace{90pt}\downarrow \\
& \color{ForestGreen}{ X^{2^*}_0 = A_{X\_X}X^1_0 + A_{X\_nbX}X^1_1 + A_{X\_\theta}\theta^1_0 + BU^1_0 }\\
& \textcolor{blue}{\phi_{X_0^2}^*(X^1_0, X^1_1, \theta^1_0, U^1_0) } = (X^{2^*}_0)^T Q_X X^{2^*}_0 \\
\end{aligned}
$
}
\end{equation}

\vspace{0.3cm}

\begin{equation}    \label{ve_1_matrix}
\resizebox{.43 \textwidth}{!} {
$
\begin{aligned}
&\begin{array} {c c c c c c} 
\color{Maroon}{W_k} & \hspace{0.3cm} \color{Maroon}{X_0^2} & \hspace{0.4cm} \color{Maroon}{X_0^1} & \hspace{0.65cm} \color{Maroon}{X_1^1} & \hspace{0.8cm} \color{Maroon}{\theta_0^1} & \hspace{0.6cm} \color{Maroon}{U_0^1} 
\end{array} \\
&\begin{bmatrix} I \\ \infty \\ \end{bmatrix}
\left[  \begin{array} {c c c c c | c}
Q_X^{\frac{1}{2}} & 0  & 0 & 0 & 0 & 0 \\
I & -A_{X\_X} & -A_{X\_nbX} & -A_{X\_\theta} & -B_X & 0 \\
\end{array} 
\right]
\Longrightarrow \\
&\begin{bmatrix} \infty \\ I \\ \end{bmatrix}
\left[ 
\begin{array} {c c c c c | c}
I & -A_{X\_X} & -A_{X\_nbX} & -A_{X\_\theta} & -B_X & 0 \\
0 & Q_X^{\frac{1}{2}}A_{X\_X} & Q_X^{\frac{1}{2}}A_{X\_nbX} & Q_X^{\frac{1}{2}}A_{X\_\theta} & Q_X^{\frac{1}{2}}B_X & 0
\end{array} 
\right]
\end{aligned}
$
}
\end{equation}

\vspace{0.3cm}

After orthogonalization, we add the second row of the factorized matrix in Eqn. \ref{ve_1_matrix} back into the factor graph as an unconstrained factor shown in blue in Fig. \ref{fig:ve_1}.

\textbf{Eliminate a pendulum node $\theta_0^2$:} Pendulum nodes are eliminated in a similar manner to cart nodes.

\begin{equation}   \label{ve_2_eqn}
\resizebox{.4 \textwidth}{!} 
{
$
\begin{aligned}
&\theta_0^{2^*} = \argminB_{\theta_0^2} \color{red} (\theta_0^2)^T Q_{\theta} \theta_0^2 \\
&\textrm{s.t.} \hspace{6pt} \textcolor{red}{ \theta^2_0 - A_{\theta\_\theta}\theta^1_0 - A_{\theta\_X} X^1_0 - A_{\theta\_nbX} X^1_1 -  B_{\theta} U_0^1 = 0} \\
&\hspace{90pt}\downarrow \\
& \color{ForestGreen}{ \theta^{2^*}_0 = A_{\theta\_\theta}\theta^1_0 + A_{\theta\_X} X^1_0 + A_{\theta\_nbX} X^1_1 + B_{\theta} U^1_0 }\\
& \textcolor{blue}{\phi_{\theta_0^2}^*(\theta^1_0, X^1_0, X^1_1, U^1_0)} = (\theta^{2^*}_0)^T Q_{\theta} \theta^{2^*}_0 \\
\end{aligned}
$
}
\end{equation}


\begin{equation}    \label{ve_2_matrix}
\resizebox{.43 \textwidth}{!} {
$
\begin{aligned}
\footnotesize 
&\begin{array} {c c c c c c} 
\color{Maroon}{W_k} & \hspace{0.3cm} \color{Maroon}{\theta_0^2} & \hspace{0.4cm} \color{Maroon}{\theta_0^1} & \hspace{0.6cm} \color{Maroon}{X_0^1} & \hspace{0.75cm} \color{Maroon}{X_1^1} & \hspace{0.5cm} \color{Maroon}{U_0^1} 
\end{array} \\
\normalsize
&\begin{bmatrix} I \\ \infty \\ \end{bmatrix}
\left[  \begin{array} {c c c c c | c}
Q_\theta^{\frac{1}{2}} & 0  & 0 & 0 & 0 & 0 \\
I & -A_{\theta\_\theta} & -A_{\theta\_X} & -A_{\theta\_nbX} & -B_\theta & 0 \\
\end{array} 
\right]
\Longrightarrow \\
&\begin{bmatrix} \infty \\ I \\ \end{bmatrix}
\left[ 
\begin{array} {c c c c c | c}
I & -A_{\theta\_\theta} & -A_{\theta\_X} & -A_{\theta\_nbX} & -B_\theta & 0 \\
0 & Q_\theta^{\frac{1}{2}}A_{\theta\_\theta} & Q_\theta^{\frac{1}{2}}A_{\theta\_X} & Q_\theta^{\frac{1}{2}}A_{\theta\_nbX} & Q_\theta^{\frac{1}{2}}B_\theta & 0
\end{array} 
\right]
\end{aligned}
$
}
\end{equation}

\vspace{0.3cm}

\textbf{Eliminate a control node $U_0^2$:} Eliminating control nodes results in a local unconstrained least squares problem, since no constrained dynamics factors lie adjacent. 

\vspace{-0.5cm}

\begin{equation}   \label{ve_3_eqn}
\resizebox{.4 \textwidth}{!} 
{
$
\begin{aligned}
&U_0^{1^*} = \argminB_{U_0^1} \{ \textcolor{red}{(U_0^1)^T Q_U U_0^1} + \textcolor{red}{\phi_{X_0^2}^*(X^1_0, X^1_1, \theta^1_0, U^1_0)} + ...\\
& \hspace{70pt} \textcolor{red}{\phi_{\theta_0^2}^*(\theta^1_0, X^1_0, X^1_1, U^1_0)} \}\\
&\hspace{90pt}\downarrow \\
& \color{ForestGreen}{ U^{1^*}_0 = -K_0^1  \left[ \begin{array}{c c c} X_0^1& X_1^1& \theta_0^1 \end{array} \right]^T }\\
&\left[ \begin{array} {c} 
\textcolor{blue}{\phi_{U_\_X}^*(X^1_0, X^1_1, \theta^1_0)} \\
\textcolor{blue}{\phi_{U_\_\theta}^*(X^1_0, X^1_1, \theta^1_0)}\\
\end{array} \right] = E_0^1 \left[ \begin{array}{c c c} X_0^1& X_1^1& \theta_0^1 \end{array} \right]^T \\
\end{aligned}
$
}
\end{equation}

\vspace{0.3 cm}

\begin{equation}    \label{ve_3_matrix}
\resizebox{.43 \textwidth}{!} {
$
\begin{aligned}
\footnotesize 
&\begin{array} {c c c c c c} 
\color{Maroon}{W_k} & \hspace{0.3cm} \color{Maroon}{U_0^1} & \hspace{0.7cm} \color{Maroon}{X_0^1} & \hspace{1.2cm} \color{Maroon}{X_1^1} & \hspace{1.1cm} \color{Maroon}{\theta_0^1}
\end{array} \\
\normalsize
&\begin{bmatrix} I \\ I \\ I \end{bmatrix}
\left[  \begin{array} {c c c c | c}
Q_X^{\frac{1}{2}}B_X & Q_X^{\frac{1}{2}}A_{X\_X} & Q_X^{\frac{1}{2}}A_{X\_nbX} & Q_X^{\frac{1}{2}}A_{X\_\theta} & 0\\
Q_\theta^{\frac{1}{2}}B_\theta &Q_\theta^{\frac{1}{2}}A_{\theta\_X} & Q_\theta^{\frac{1}{2}}A_{\theta\_nbX} &Q_\theta^{\frac{1}{2}}A_{\theta\_\theta} & 0\\
Q_U ^{\frac{1}{2}} & 0 & 0 & 0 & 0 \\
\end{array} 
\right] \Longrightarrow \\
& \left[ \arraycolsep=0.1pt\def\arraystretch{1.5}
\begin{array} {c} 
R_0^1 \\ \begin{bmatrix} I \\ I \end{bmatrix} \\ \end{array} 
\right]
\left[ \arraycolsep=7pt\def\arraystretch{1.5}
\begin{array} {c c | c}
I &  \begin{bmatrix} K_0^1 \end{bmatrix} & 0 \\
\begin{bmatrix} 0\\0 \end{bmatrix} & \begin{bmatrix} E_0^1 \end{bmatrix} & \begin{bmatrix} 0\\0 \end{bmatrix}
\end{array} 
\right] \\
\end{aligned}
$
}
\end{equation}
where $R_0^1$, $E_0^1$ and $K_0^1$ are from QR factorization of $F_k$. This yields a feedback policy for $U_0^1$ (in green) and cost-to-go values $\phi$ (in blue). Note that the last step of Eqn. \ref{ve_3_matrix} lumps together the columns corresponding to $\{X_0^1, X_1^1, \theta_0^1\}$.

\begin{align}
F_k & \rightarrow Q \left[ \begin{array}{c c} R_0^1 & T_0^1 \\ 0 & E_0^1 \end{array} \right]\\
K_0^1 & = -(R_0^1)^{-1} T_0^1
\end{align}

\vspace{0.3 cm}


\subsection{Variable Ordering} \label{ordering}
The run time of each local sub-problems is determined by the size of $F_k$ matrix. The $F_k$ matrix has dimensions of $p_1 n \times p_2 n$ where $p_1$ is the number of factors attached to the node in question, $p_2$ is the number of variables in the separator, and $n$ is the state-space dimension of each pendulum/cart node (equals 2 in our example). We expect that for all $F_k$, $p_1$ and $p_2$ are bounded constant. While Section \ref{ve_details} describes a choice of elimination ordering that is intuitive and easy to follow, it creates unnecessary non-zero elements (\textit{fill-in}) when we factorize each $F_k$ matrix, which is equivalent to adding more factors to the factor graph. Then subsequent local sub-problems will have larger dimensions because the separator contains more nodes. Different elimination ordering results in different sub-problem size affecting run time performance. 

Since choosing an optimal elimination order is long known to be NP-complete \cite{yannakakis1981computing}, in the numerical analysis community, various heuristic-based methods were developed \cite{amestoy1996approximate, demmel1999supernodal, davis2004column}. 
Following \cite{DBLP:journals/corr/abs-1911-10065}, we use Column Approximate Minimum Degree (ColAMD) \cite{davis2004column}. ColAMD selects variable order according to a metric that evaluates potential fill-in introduced when a variable is eliminated. 


\subsection{Complexity Analysis} \label{complexity}
Given a proper elimination order, it follows from the structural sparsity that $p_1$ and $p_2$ are constants as the state and control spaces scale. Then the total complexity of our method is $O((M+N_s)T)$ compared to cubic run time complexity in state space size of conventional DP approaches for LQR. 


The speed-up comes from breaking the original problem into $(M+N_s)T$ local sub-problems. While eliminating cart nodes and pendulum nodes, a constrained least-squares problem is solved using MGS with $O(p_1n (p_2n)^2)$ floating point operations \cite{gulliksson1995modified}. When we eliminate control nodes, we are solving a local unconstrained least-squares problem to factorize $F_k$ using QR factorization, which in the worst case requires $O(p_1n (p_2n)^2)$ steps \cite{trefethen1997numerical}. This is followed by back-substitution which requires $O((p_1n)^2+(p_2n)^2)$ flops. Hence, the complexity of each local sub-problem is constant. 

Finally, the ColAMD algorithm has complexity $O(m)$ for a bounded degree graph with $m$ edges \cite{amestoy1996approximate}, which is $O((M+N_s)T)$ in our case. Moreover, the ordering can be precomputed offline if the system structure and time horizon are fixed. The relationship between the link connection structure and the optimal elimination order is an interesting avenue for future work, since the optimal elimination order can further speed up the factor graph optimization.

\section{Results}
In this section, we evaluate SGOPT and present a comparison with conventional LQR methods. Most of the analysis is presented for the linear example of stabilization about the unstable equilibrium, except for the last section which addresses extensions for solving the nonlinear swing-up problem. SGOPT is implemented using the GTSAM \cite{dellaert2012factor} library in C++. The first step is to validate the SGOPT solution by comparing it with the baselines on a relatively small-scale problem. Next, we scale the system dimensionality to demonstrate superior runtimes and numerical conditioning as compared to the conventional LQR solvers.
\subsection{Implementation Details}
We benchmark our results with two baseline methods. The first is the C++ based LQR solver in the Control Toolbox library \cite{adrlCT}. The second is our independent DP implementation of LQR in C++ that relies on column-pivot Householder QR factorization implementation in the Eigen library \cite{eigenweb} for solving the Riccati equations at each time step. The parameters used in our experiments are 
$m_c=1kg$, $m_p=0.2kg$, $\ell=0.5m$, $Q_X=Q_\theta=10I_{2\times2}$, $Q_{f_X}=Q_{f_\theta}=3000I_{2\times2}$, $Q_U=0.01$, $\Delta t=0.05s$, $k_{spring}=1000\frac{N}{m}$, $c_{damping}=1\frac{Ns}{m}$.


\subsection{Validation Analysis}
\begin{figure}
  \centering
  \includegraphics[width=0.42\textwidth]{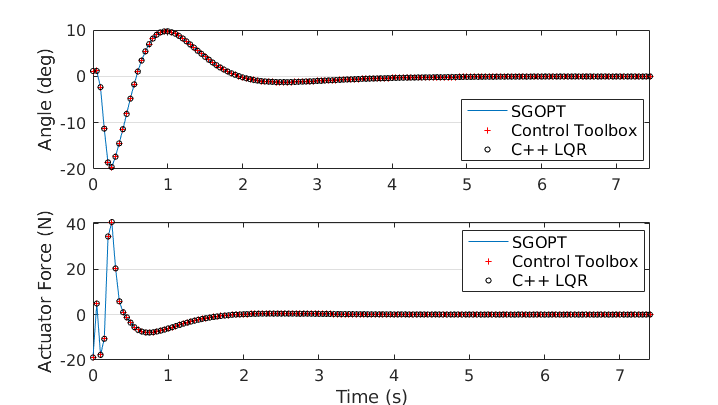}
  \caption{\footnotesize{(Top) $\theta$ trajectory of the pendulum on the underactuated cart; (Bottom) Control trajectory for the first actuated cart.}}  
  \label{fig:time_resp}
\end{figure}

We first compare the three methods on a relatively small-scale problem to validate the solution equivalence of SGOPT to conventional LQR solvers. We consider the example of a 3-CP system with two actuated carts and extract the time response for 150 time-steps given an initial perturbation offset of $\theta_0 = 1.15$ deg from vertical for all pendulums. Runtime and total cost are tabulated in Table \ref{tab:lqr_timing}, while Fig \ref{fig:time_resp} plots the time-response. We observe very similar cost and time responses from all solvers, while the runtime for SGOPT is around 3x slower. This is because the GTSAM backend implementation for constrained QR factorization is currently un-optimized, and hence adds a compute overhead for smaller systems. However, as we will demonstrate in the next section, this overhead becomes insignificant as the system scales due to linear vs cubic complexity.

\begin{table}[ht!]
  \begin{center}
    \caption{Time-response results}
    \begin{tabular}{lcr}
        \toprule
        \textbf{Method} & \textbf{Cost} & \textbf{Run-time (s)}\\
        \midrule 
        SGOPT & 2011.26 & 0.0687 \\
        Control Toolbox & 2011.26 & 0.0226 \\
        C++ LQR & 2011.29 & 0.0252 \\
        \bottomrule
    \end{tabular}
    \label{tab:lqr_timing}
  \end{center}
\end{table}

\subsection{Runtime and Cost Analysis}
In this section, we demonstrate linear growth in runtime for SGOPT as the state and control dimensions scale when compared to the benchmark DP solvers, for a structurally sparse system. Additionally, we can also observe superior numerical conditioning compared to the DP approach as the dynamical system becomes more complex, which manifests in lower total cost solutions. This is due to the VE procedure resulting in far simpler local inversions on sub-matrices as opposed to dense inversions on large matrices at each DP step. 

\begin{figure} 
\centering
\begin{subfigure}{.24\textwidth}
  \centering
  \includegraphics[width=1.05\linewidth, trim={0.2cm 0.1cm 0.2cm 0.5cm},clip]{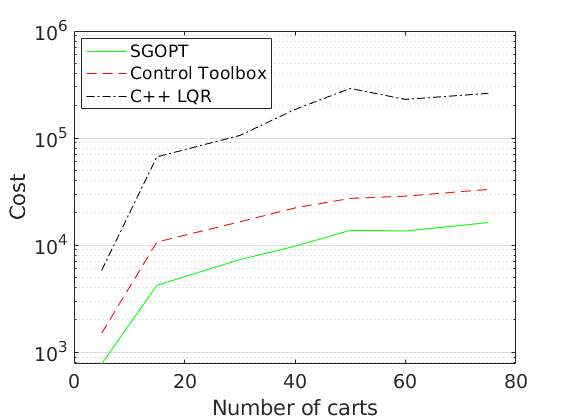}
  \label{fig:sub1}
\end{subfigure}%
\hfill
\begin{subfigure}{.24\textwidth}
  \centering
  \includegraphics[width=1.05\linewidth, trim={0.2cm 0.1cm 0.2cm 0.5cm},clip]{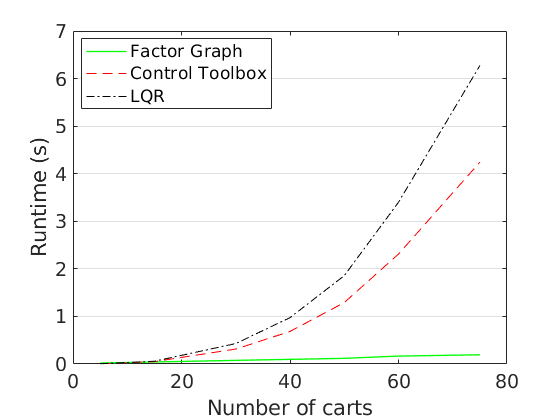}
  \label{fig:sub2}
\end{subfigure}%
\vspace{-0.2cm}
\begin{subfigure}{.24\textwidth}
  \centering
  \includegraphics[width=1.05\linewidth, trim={0.2cm 0.1cm 0.2cm 0.5cm},clip]{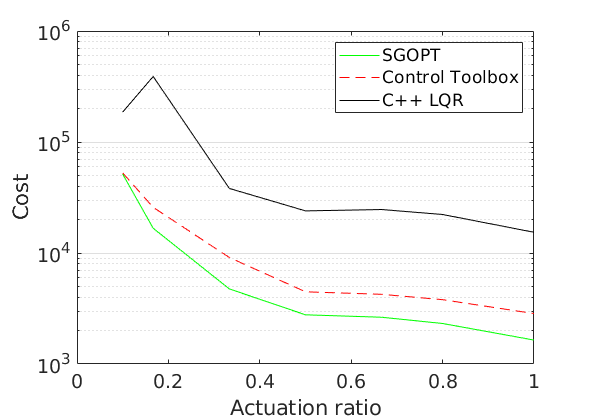}
\end{subfigure}%
\hfill
\begin{subfigure}{.24\textwidth}
  \centering
  \includegraphics[width=1.05\linewidth, trim={0.2cm 0.1cm 0.2cm 0.5cm},clip]{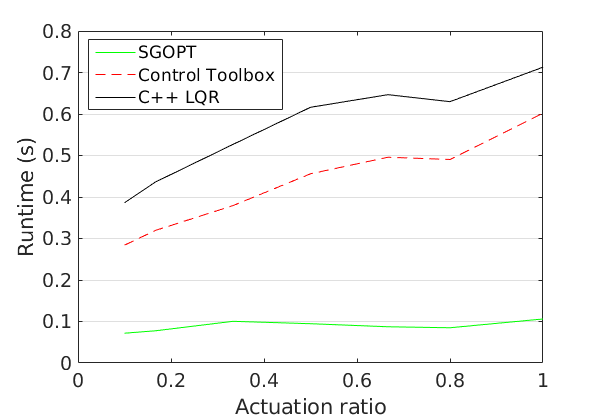}
\end{subfigure}
\caption{ \footnotesize{Runtime and Cost comparison for $T$=10 steps as $N$ scales for constant $\rho$ = 0.25 (top); as $\rho$ scales for constant $N$=30 (bottom)}}
\label{fig:scalability}
\end{figure}

The top part of Fig \ref{fig:scalability} shows runtime and cost as we scale the number of carts while keeping actuation ratio ($\rho$) constant at 0.25. It is evident that OCP solutions using DP approaches face a computational bottleneck in the state dimensionality, which is not the case with SGOPT. The bottom part of Fig \ref{fig:scalability} similarly looks at varying $\rho$ for $N$ held at 30 over 10 time-steps, and here too we can observe slower growth in runtime as control dimension increases compared to DP solvers. 

\subsection{Extension to Nonlinear Swing Up}
For solving the swing-up problem, we iteratively solve SGOPT by linearizing about a trajectory (starting with an initial guess) until convergence, similar to standard iterative optimization methods such as SQP \cite{gill2005snopt} and DDP \cite{mayne1973differential}. The inner linear optimization is solved using SGOPT and outer loop is governed by the Levenberg-Marquardt algorithm for regularization.

\begin{figure}
  \centering
  \includegraphics[width=0.4\textwidth, trim={0.5cm 0 0.5cm 0},clip]{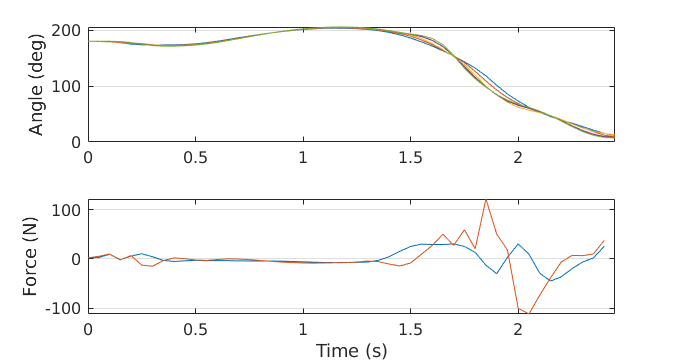}
  \caption{\footnotesize{Angles of the 5 pendulums (top) and forces applied by the 2 actuators (bottom) over time for the swing-up solution. Cost = 34719, Run time = 6.15s, Time horizon = 50 steps, Iterations till convergence = 38.}}
  \label{fig:swingup}
\end{figure}

Results are presented for iterative-SGOPT solving the swing-up problem for $N=5$, and $M=2$ starting from stable equilibrium, with a time-response shown in Fig. \ref{fig:swingup}. A comparison with benchmark solvers is not as straightforward in the nonlinear case since there are multiple factors governing convergence speed, regularization, behavior around local minima, etc. Preliminary results from iterative-SGOPT are displayed here as a demonstration of solver capability, while the rigorous characterization and comparison with benchmark solvers is left for future work.

\vspace{-0.2cm}

\section{Conclusion}
In this paper we proposed an intuitive modeling paradigm that uses factor graphs to capture the structural sparsity pattern of the OCP for multi-rigid-body systems. We also presented the Variable Elimination algorithm, that capitalizes on sparsity in the factor graph to compute efficient and scalable solutions to the OCP. We utilized the N-cartpole system as a canonical example to demonstrate the inner workings of the modeling and inference, for which we presented benchmark experiments to validate the effectiveness of our approach. Our approach can be extended to control more complicated nonlinear multi-rigid-body systems.

\addtolength{\textheight}{-12cm}   









\medskip

\bibliographystyle{unsrt}
\bibliography{IEEEexample}

\begin{thebibliography}{10}

\bibitem{boyd2004convex}
Stephen Boyd and Lieven Vandenberghe.
\newblock {\em Convex optimization}.
\newblock Cambridge university press, 2004.

\bibitem{Laine2019EfficientCO}
F.~Laine and C.~Tomlin.
\newblock Efficient computation of feedback control for equality-constrained
  lqr.
\newblock {\em 2019 International Conference on Robotics and Automation
  (ICRA)}, pages 6748--6754, 2019.

\bibitem{Kaess-2017-106169}
Frank Dellaert and Michael Kaess.
\newblock {\em Factor Graphs for Robot Perception}.
\newblock Now Publishers Inc., August 2017.

\bibitem{5979641}
M.~{Kaess}, H.~{Johannsson}, R.~{Roberts}, V.~{Ila}, J.~{Leonard}, and
  F.~{Dellaert}.
\newblock isam2: Incremental smoothing and mapping with fluid relinearization
  and incremental variable reordering.
\newblock In {\em 2011 IEEE International Conference on Robotics and
  Automation}, pages 3281--3288, 2011.

\bibitem{gill2005snopt}
Philip~E Gill, Walter Murray, and Michael~A Saunders.
\newblock Snopt: An sqp algorithm for large-scale constrained optimization.
\newblock {\em SIAM review}, 47(1):99--131, 2005.

\bibitem{baraff1996linear}
David Baraff.
\newblock Linear-time dynamics using lagrange multipliers.
\newblock In {\em Proceedings of the 23rd annual conference on Computer
  graphics and interactive techniques}, pages 137--146, 1996.

\bibitem{brudigam2020linear}
J~Br{\"u}digam and Z~Manchester.
\newblock Linear-time variational integrators in maximal coordinates.
\newblock In {\em Workshop on the Algorithmic Foundations of Robotics (WAFR)},
  2020.

\bibitem{featherstone2014rigid}
Roy Featherstone.
\newblock {\em Rigid body dynamics algorithms}.
\newblock Springer, 2014.

\bibitem{10.1145/1553374.1553508}
Marc Toussaint.
\newblock Robot trajectory optimization using approximate inference.
\newblock In {\em Proceedings of the 26th Annual International Conference on
  Machine Learning}, ICML '09, page 1049–1056, New York, NY, USA, 2009.
  Association for Computing Machinery.

\bibitem{Mukadam_2018}
Mustafa Mukadam, Jing Dong, Xinyan Yan, Frank Dellaert, and Byron Boots.
\newblock Continuous-time gaussian process motion planning via probabilistic
  inference.
\newblock {\em The International Journal of Robotics Research},
  37(11):1319–1340, Sep 2018.

\bibitem{dong2016motion}
Jing Dong, Mustafa Mukadam, Frank Dellaert, and Byron Boots.
\newblock Motion planning as probabilistic inference using gaussian processes
  and factor graphs.

\bibitem{DBLP:journals/corr/abs-1911-10065}
Mandy Xie and Frank Dellaert.
\newblock A unified method for solving inverse, forward, and hybrid manipulator
  dynamics using factor graphs.
\newblock {\em arXiv preprint arXiv:1911.10065}, 2019.

\bibitem{xie2020factorgraph}
Mandy Xie, Alejandro Escontrela, and Frank Dellaert.
\newblock A factor-graph approach for optimization problems with dynamics
  constraints, 2020.

\bibitem{xie2020batch}
Mandy Xie and Frank Dellaert.
\newblock Batch and incremental kinodynamic motion planning using dynamic
  factor graphs.
\newblock {\em arXiv preprint arXiv:2005.12514}, 2020.

\bibitem{9196649}
P.~{Sodhi}, S.~{Choudhury}, J.~G. {Mangelson}, and M.~{Kaess}.
\newblock Ics: Incremental constrained smoothing for state estimation.
\newblock In {\em 2020 IEEE International Conference on Robotics and Automation
  (ICRA)}, pages 279--285, 2020.

\bibitem{5652875}
A.~{Cunningham}, M.~{Paluri}, and F.~{Dellaert}.
\newblock Ddf-sam: Fully distributed slam using constrained factor graphs.
\newblock In {\em 2010 IEEE/RSJ International Conference on Intelligent Robots
  and Systems}, pages 3025--3030, 2010.

\bibitem{6842254}
D.~{Ta}, M.~{Kobilarov}, and F.~{Dellaert}.
\newblock A factor graph approach to estimation and model predictive control on
  unmanned aerial vehicles.
\newblock In {\em 2014 International Conference on Unmanned Aircraft Systems
  (ICUAS)}, pages 181--188, 2014.

\bibitem{Rodrguez2017AFG}
J.~Rodr{\'i}guez and Iv{\'a}n Dar{\'i}o.
\newblock A factor graph approach to constrained optimization.
\newblock 2017.

\bibitem{yang2020equality}
Shuo Yang, Gerry Chen, Yetong Zhang, Frank Dellaert, and Howie Choset.
\newblock Equality constrained linear optimal control with factor graphs.
\newblock In {\em IEEE International Conference on Robotics and Automation
  (ICRA)}, 2021.

\bibitem{gulliksson1995modified}
M{\aa}rten Gulliksson.
\newblock On the modified gram-schmidt algorithm for weighted and constrained
  linear least squares problems.
\newblock {\em BIT Numerical Mathematics}, 35(4):453--468, 1995.

\bibitem{yannakakis1981computing}
Mihalis Yannakakis.
\newblock Computing the minimum fill-in is np-complete.
\newblock {\em SIAM Journal on Algebraic Discrete Methods}, 2(1):77--79, 1981.

\bibitem{amestoy1996approximate}
Patrick~R Amestoy, Timothy~A Davis, and Iain~S Duff.
\newblock An approximate minimum degree ordering algorithm.
\newblock {\em SIAM Journal on Matrix Analysis and Applications},
  17(4):886--905, 1996.

\bibitem{demmel1999supernodal}
James~W Demmel, Stanley~C Eisenstat, John~R Gilbert, Xiaoye~S Li, and Joseph~WH
  Liu.
\newblock A supernodal approach to sparse partial pivoting.
\newblock {\em SIAM Journal on Matrix Analysis and Applications},
  20(3):720--755, 1999.

\bibitem{davis2004column}
Timothy~A Davis, John~R Gilbert, Stefan~I Larimore, and Esmond~G Ng.
\newblock A column approximate minimum degree ordering algorithm.
\newblock {\em ACM Transactions on Mathematical Software (TOMS)},
  30(3):353--376, 2004.

\bibitem{trefethen1997numerical}
Lloyd~N Trefethen and David Bau~III.
\newblock {\em Numerical linear algebra}, volume~50.
\newblock Siam, 1997.

\bibitem{dellaert2012factor}
Frank Dellaert.
\newblock Factor graphs and {GTSAM}: A hands-on introduction.
\newblock Technical report, Georgia Institute of Technology, 2012.

\bibitem{adrlCT}
Markus Giftthaler, Michael Neunert, Markus St{\"a}uble, and Jonas Buchli.
\newblock The control toolbox — an open-source c++ library for robotics,
  optimal and model predictive control.
\newblock {\em 2018 IEEE International Conference on Simulation, Modeling, and
  Programming for Autonomous Robots (SIMPAR)}, pages 123--129, 2018.

\bibitem{eigenweb}
Ga\"{e}l Guennebaud, Beno\^{i}t Jacob, et~al.
\newblock Eigen v3.
\newblock http://eigen.tuxfamily.org, 2010.

\bibitem{mayne1973differential}
David~Q Mayne.
\newblock Differential dynamic programming--a unified approach to the
  optimization of dynamic systems.
\newblock In {\em Control and Dynamic Systems}, volume~10, pages 179--254.
  Elsevier, 1973.

\end{thebibliography}

\end{document}